\documentclass{article}

\usepackage[utf8]{inputenc}

\usepackage{microtype}
\usepackage{graphicx}
\usepackage{tabularx}
\usepackage{gensymb}
\usepackage{booktabs}
\usepackage{hyperref}
\usepackage{subfig}

\usepackage[accepted]{icml2019}

\begin{document}

\twocolumn[
\icmltitle{Increasing Shape Bias in ImageNet-Trained Networks Using Transfer Learning and Domain-Adversarial Methods}

\begin{icmlauthorlist}
\icmlauthor{Francis Brochu}{ul}
\end{icmlauthorlist}

\icmlaffiliation{ul}{Department of Computer Science and Software Engineering, Université Laval, Québec, Québec, Canada}

\icmlcorrespondingauthor{Francis Brochu}{francis.brochu.2@ulaval.ca}

\icmlkeywords{Machine Learning, Deep Learning, Computer Vision}

\vskip 0.3in
]

\printAffiliationsAndNotice{}

\begin{abstract}

Convolutional Neural Networks (CNNs) have become the state-of-the-art method to learn from image data.
However, recent research shows that they may include a texture and colour bias in their representation, contrary to the intuition that they learn the shapes of the image content and to human biological learning.
Thus, recent works have attempted to increase the shape bias in CNNs in order to train more robust and accurate networks on tasks.
One such approach uses style-transfer in order to remove texture clues from the data.
This work reproduces this methodology on four image classification datasets, as well as extends the method to use domain-adversarial training in order to further increase the shape bias in the learned representation.
The results show the proposed method increases the robustness and shape bias of the CNNs, while it does not provide a gain in accuracy.

\end{abstract}

\section{Introduction}

Convolutional Neural Networks (CNNs) have become, in the last few years, the state-of-the-art method to classify, segment or otherwise learn from the content of images. \cite{rawat2017deep}.
The rise of CNNs goes back to the ImageNet competition of 2012 \cite{krizhevsky2012imagenet}.
They have then maintained their domination of that competition in the following years over other types of methods \cite{szegedy2015going} \cite{resnet}. 
A common explanation for the workings of a CNN is that the successive layers of convolutions and pooling learn to represent the edges and shapes within the images, and assemble these representations into more complicated shapes that mirror the problems \cite{kriegeskorte2015deep}.

However, in recent years, some results have come to challenge this assumption.
Some articles posit that CNNs can still classify images where any object shape is absent, and where texture information is still present \cite{gatys2017texture} \cite{brendel2019approximating}.
Moreover, many architectures have difficulty with sketches, where shape information is clearly present and there are no texture clues \cite{ballester2016performance}.

\begin{figure}
    \centering
    \includegraphics[width=8cm]{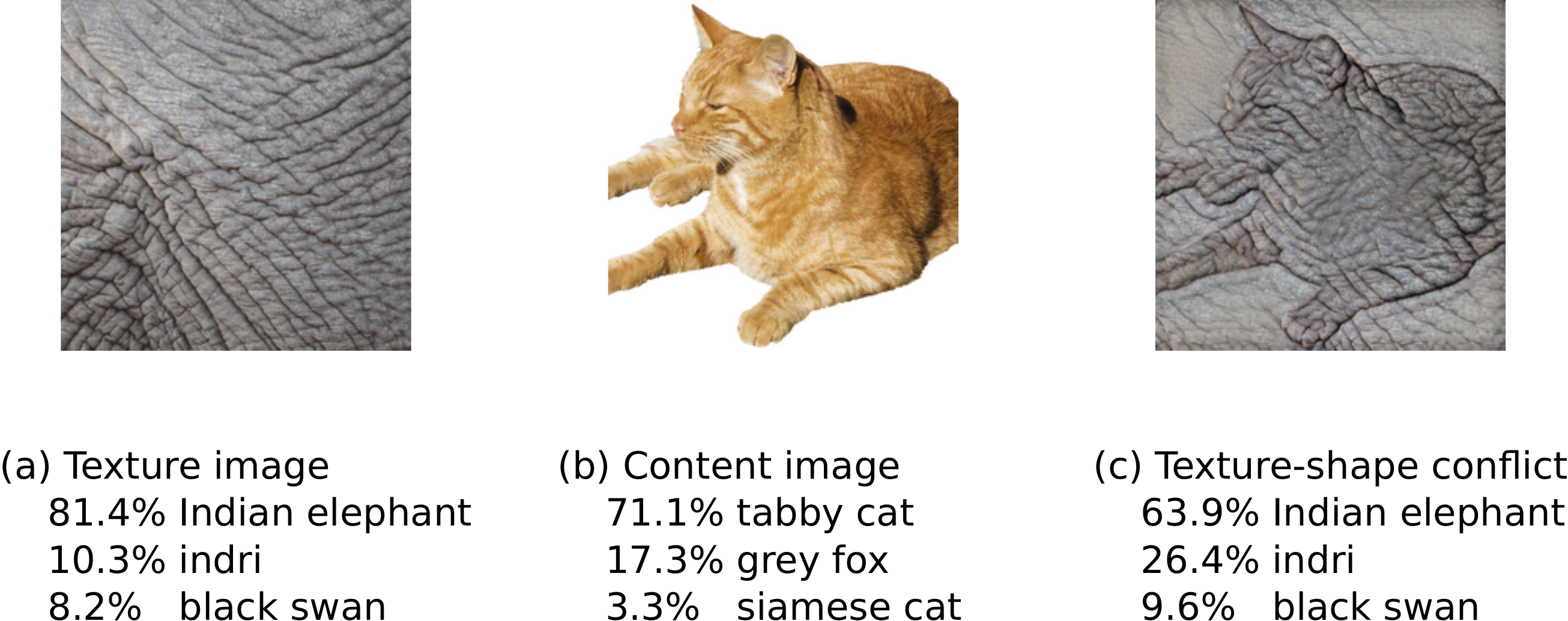}
    \caption{Example of texture and shape clues conflict \cite{geirhos2018}}
    \label{fig:texture_cat}
\end{figure}

While many other results also exist to support that CNNs do indeed learn shape information, there is a question to be raised as to whether CNNs learn a mixture of the texture information and shape information \cite{ritter2017cognitive}.
It is known that humans learn mainly by shape information \cite{ritter2017cognitive}.
Thus, \cite{geirhos2018} explores this problem by comparing human experimentation and CNNs with images with missing or changed texture clues conflicting with shape information, as shown in Figure~\ref{fig:texture_cat}.
These results show that CNNs pretrained on ImageNet have learned from texture and colour information.
A workflow to increase the shape-bias of CNNs on ImageNet is then presented, using mainly style-transfer \cite{adaIN}.
The article concludes that this workflow increases both the robustness and accuracy of CNNs.

\section{Material and Methods}

The following section will introduce and describe the different datasets used for the experiments.
The network architectures used will also be described.
Then, the methodology used to reproduce \cite{geirhos2018} will be described.
Finally, the domain-adversarial training of neural networks will be detailed.

\subsection{Datasets}

Four image datasets were chosen to conduct the experiments.
For each of these datasets, there is an associated classification task.
Thus, the basic same architectures can be applied to each dataset, with only the number of classes varying between them.
Each classification task intuitively varies in difficulty.
Let us now consider each dataset in turn, from the easiest to hardest tasks.

The first is the Dice dataset, taken from Kaggle Datasets \cite{DiceData}.
The dataset is composed of 16 000 images of polyhedral dice.
There are thus 6 classes in this dataset: those of d4's, d6's, d8's, d10's, d12's and d20's, where the number describes the number of faces on the die.
The dataset itself is relatively simple, as it was acquired in very controlled circumstances.
The dataset was generated by rotating video cameras around the dice on a table, at varying angles and in front of 6 different backgrounds (white, coloured and wood-textured).
Most images were then cropped to $480 \times 480$ pixels, and any image where the die is partially or completely out of crop were removed.
Some images of d6's and d20's are $1024 \times 1024$, as they were from an earlier version of the dataset.

The second dataset was the Dogs versus Cats (DvC) dataset, taken from the Kaggle competition \cite{DvCData}.
This dataset is formed of 25 000 labelled images of cats and dogs.
The classification task is thus binary in this case, predicting whether the image contains a dog or a cat.
Most images in this dataset were of a size greater than $224 \times 224$, but some were shorter along either axis.

These first two tasks are intuitively relatively easy.
The Dice dataset is a relatively artificial dataset, as even if its images are real images, they are generated in order to form a dataset and are without noise.
Each image is correctly classified, there are no partially cropped or missing dice, and each image contains a single die.
In addition, it is an easy task using only shape and a human would have a perfect or almost perfect accuracy on the task.
The second task is formed of more real-world images, but it is still a relatively simple task which would be easy for a person.
Also, we can easily imagine that shape is important, and perhaps all that is necessary, in order to spot whether an animal is a dog or cat.
Still, some images contain multiple animals (of the same type), or a human in addition to the animals, and a wide variety of backgrounds.

The third dataset is the Dog Breed Identification (DBI) playground competition from Kaggle \cite{DBIData}.
This dataset is formed of 10 223 images of 120 different breeds of dogs.
The classification task for this dataset is thus to predict which of the 120 classes the image belongs to.
With the relatively low number of images (the lowest of the datasets), and the high number of classes, this classification task is much more difficult than the previous two.
Moreover, some of the dog breeds included in the classes are difficult to predict without error, even for humans.
For example, two of the classes are miniature schnauzers and standard schnauzers, which look practically identical, except for size.
Thus, a lower accuracy is expected on this task.
Another measure than a simple accuracy will be calculated on this dataset for the results.

The final dataset used is the Food101 dataset, found on Kaggle Datasets \cite{Food101Data}.
This dataset is formed of 101 types of food, with 1000 images per class.
Thus, the total dataset is composed of 101 000 images.
As with the DBI dataset, another measure than accuracy will be used for the predictions on this dataset, due to the number of classes.
Note than multiple classes in this dataset are close and hard to classify, as there are, for example, 5 different types of soup in the dataset.

\subsubsection{Splitting the datasets}

Each dataset received the same treatment in order to be split into training, validation and test sets.
First, if the Kaggle dataset was already split into training and validation sets, they were merged back into a single dataset.
Then, 20\% of the images in each class were randomly selected to form the test set.
Afterwards, a further 20\% of the remaining samples were randomly selected to form the validation set.
The validation set is thus 16\% of the total number of images per class.
The remaining images (64\% per class) form the training set.

\subsection{Model Architectures}

Two model architectures were chosen for the experiments.
The first is the ResNet 34-layer (ResNet-34) architecture \cite{resnet}.
This model contains 34 layers, from a $7 \times 7$ convolution layer and max pooling at the base of the network, and using blocks of $3 \times 3$ convolutions layers, with residual links every 2 layers.
The number of channels increase as the layers near the end of the network.
Finally, the classification layers are formed by an average pooling and a fully connected layer.
The architecture reports a 8.58\% Top-5 error on ImageNet.
The computational demand per image is of 3.6 GFLOPs for this network.

The second chosen architecture is the SqueezeNet (1.1) architecture \cite{squeezenet}.
This model reports AlexNet-level accuracy on the ImageNet dataset, at 19.38\% Top-5 error.
However, this architecture is designed to contain much less parameters, and thus can be trained and can predict much faster than AlexNet.
The article reports SqueezeNet 1.0 as 512x smaller than AlexNet.
SqueezeNet 1.1 reports the same accuracy as SqueezeNet 1.0, but with a further 2.4 times less calculations, for a total of 0.72 GFLOPs per image.

For the experiments, the pretrained networks were implemented and executed in PyTorch \cite{pytorch}.
The final pretrained fully connected layer of ResNet-34 was replaced by a randomly initialized fully connected layer in order to adapt it to the number of classes in the given dataset (2, 6, 101 or 120).
This random initialization was the default PyTorch initialization.
In the case of SqueezeNet 1.1, the number of Global Average Pools was changed from 1000 (for ImageNet) to the number of classes in the dataset. 
These two architectures were chosen for their relatively small sizes and fast computations.

\subsection{Applying Geirhos' methodology to transfer learning}

Firstly, let us consider the reproduction of the results in \cite{geirhos2018}.
Each dataset will be style-transferred using the same AdaIN architecture \cite{adaIN} and dataset as indicated in the article.
Moreover, this can be easily done as the author have published a git repository containing all necessary code in order to apply this style transfer methodology \cite{stylized-imagenet}.
Note that the images generated by this method are already cropped and resized to $224 \times 224$, as the AdaIN architecture uses a VGG model as an encoder for the images.

Thus, each of the four datasets (Dice, DvC, DBI and Food101) were style-transferred using this code, and the Paint-by-numbers Kaggle competition dataset, as described by Geirhos \textit{et al.}.
We thus obtain two versions of the datasets.
The first is the original dataset, which be henceforth be referenced as the \textbf{base} dataset.
The second is the style-transferred dataset, which will be reffered as the \textbf{stylized} dataset.
The combined dataset formed by the union of the base and stylized dataset is known as the \textbf{mixed} dataset.

\begin{figure*}[h]
    \centering
    \includegraphics[width=\textwidth]{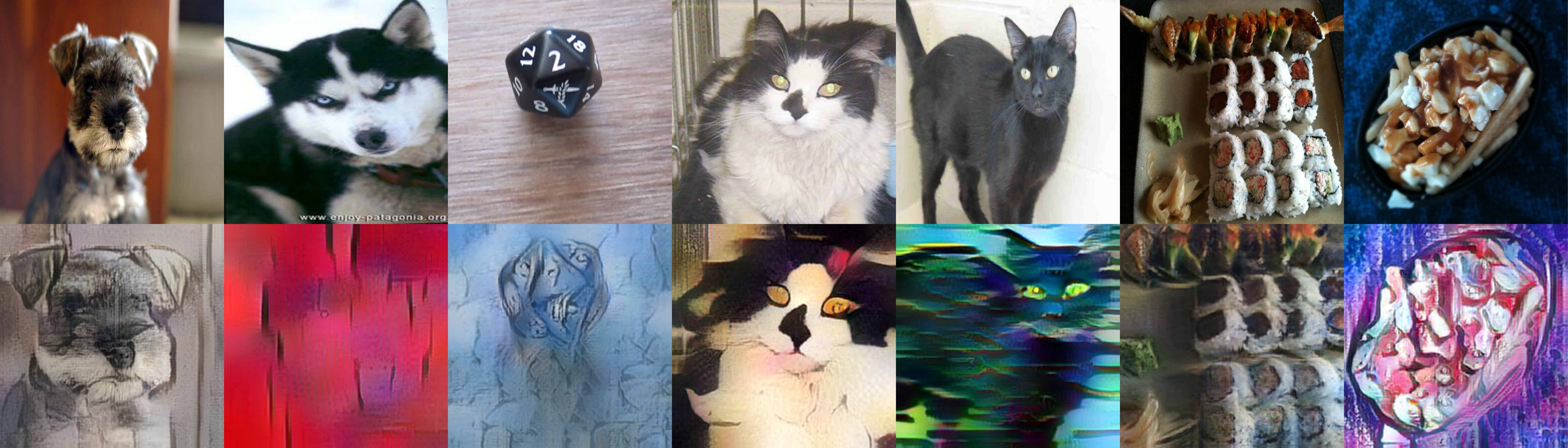}
    \caption{Some sample images from the 4 datasets, and their style-transferred equivalents.}
    \label{fig:ex_style}
\end{figure*}

The figure~\ref{fig:ex_style} shows some examples from the chosen datasets of base images and their stylized counterparts.
We can see that multiple images are still easy to recognize and classify, but others will undergo more radical transformations and may end up being much harder to predict or classify.
The second image from the left in the figure shows the latter case.

\subsubsection{Training the models}

The code used to train the models is available on \hyperlink{https://github.com/francisbrochu/DAStylizedTraining}{Github}.
Each model, for each experiment type (base, stylized and mixed) was trained in a similar manner.

First, the images were loaded in batches of 64, randomly shuffled for each epoch.
After some early experimentation (results not shown), the data augmentation was fixed for each dataset.
The details of that data augmentation is shown in the next subsection.

As stated before, the models were implemented in PyTorch, and their pretrained weights on ImageNet were set using torchvision (except the classification layer).
Since the classification layer was not pretrained, it could have a specific training rate specified, different that the rest of the dataset.
All experiments were conducted using the Adam optimizer implemented by PyTorch \cite{adam}.
Some combinations of learning rates (classification learning rate, general learning rate) were tested initially on each dataset, for each experiment.
When the optimal result (on the validation set) was one of the boundaries of the search space, the search space was expanded by an order of magnitude.
If not, some combinations between the lowest ones observed were tested.
Those default combinations were $(10^{-3}, 10^{-3})$, $(10^{-3}, 10^{-4})$ and $(10^{-4}, 10^{-4})$.
Generally, the $(10^{-3}, 10^{-4})$ combination was the best one, as observed on the validation set.

Additionally, a Multi-Step Learning Rate scheduler was used.
The milestones for the scheduler were added manually, by observing the training losses and accuracies.
The step scheduler was tested with a gamma of $0.2$ and $0.5$.
Note that most models were trained with a gamma of $0.5$ and multiple steps, as it seemed to yield lower error rates.

Another parameter that was explored was the weight decay (or L2 regularization).
Different values were tested between $10^{-3}$ and $10^{-5}$, with the search space being extended if the optimal parameter value was found to be at either edge of the space.
Most experiments (base, style and mixed) tended to adopt the same or similar learning rates on the same dataset.

Finally, early stopping was added to the training.
This ended the training and saved the best model observed (on the validation set) if not model performed better for a number of epochs equal to a specified patience.
Depending on the experiment and dataset, this patience varied mostly within the 20 to 30 range, with some experiments going as far as 40 or 50 epochs.

\subsubsection{Data augmentation}

Firstly, let us consider the data augmentation on images from the base datasets.
Every image had a 50\% chance of being flipped horizontally, in every dataset.
Then, a random choice was introduced in order to obtain a $224 \times 224$ pixel image.
In all cases, an image could simply be resized to $224 \times 224$.
Alternatively, a random crop of $224 \times 224$ could be applied, with zero-padding added if necessary.
The image could also be treated through a a random resized crop (as defined by the torchvision library).
This random resized crop was to crop to $224 \times 224$ pixels, and would choose a random proportion of the image.
The precise proportion is variable by dataset.
In the DBI dataset, the minimal proportion to be kept was 50\%, and the maximal 100\%.
In the other datasets, the minimal proportion was of 33\%, and the maximum 100\%.

The Dice dataset was the only dataset where no image was smaller than $224 \times 224$ pixels.
Thus, an additional choice was possible in the data augmentation.
The image could undergo a random rotation of up to $\pm 45^{\circ}$, followed by a center crop of size $224 \times 224$.

As the stylized datasets were already cropped to $224 \times 224$ pixels by the style-transfer network, they underwent a reduced data augmentation.
The stylized dataset could then simply be left at their normal size, or undergo a random resized crop of between 50\% and 100\% of the image.

Finally, before image resizing or cropping, all the images in the Dice and DvC datasets received a random color jitter, of up to 25\% of brightness, hue, saturation and contrast.
This color jitter was applied in both the base and stylized datasets.

For the mixed experiments, the data augmentation of the base images is the same as the base dataset, and that of the stylized images the same as the stylized dataset.

\subsection{Domain-Adversarial Training of Networks}

A further method to increase shape-bias in the learned representation can be achieved using \emph{domain-adversarial} training \cite{DANN}.
This type of neural network architecture and training aims to learn a representation that can classify the classes within the dataset, while also learning not to distinguish two \emph{domains}.
In the case of the original Domain-Adversarial Neural Network (DANN) article, this took the form of transfer learning and source and target domains.
The network trained the classification head of the network on the source domain, while it also trained a domain-adversarial head using labelled source and target domains.
The final model was then used to predict on the target domain.

\begin{figure}[h]
    \centering
    \includegraphics[width=8cm]{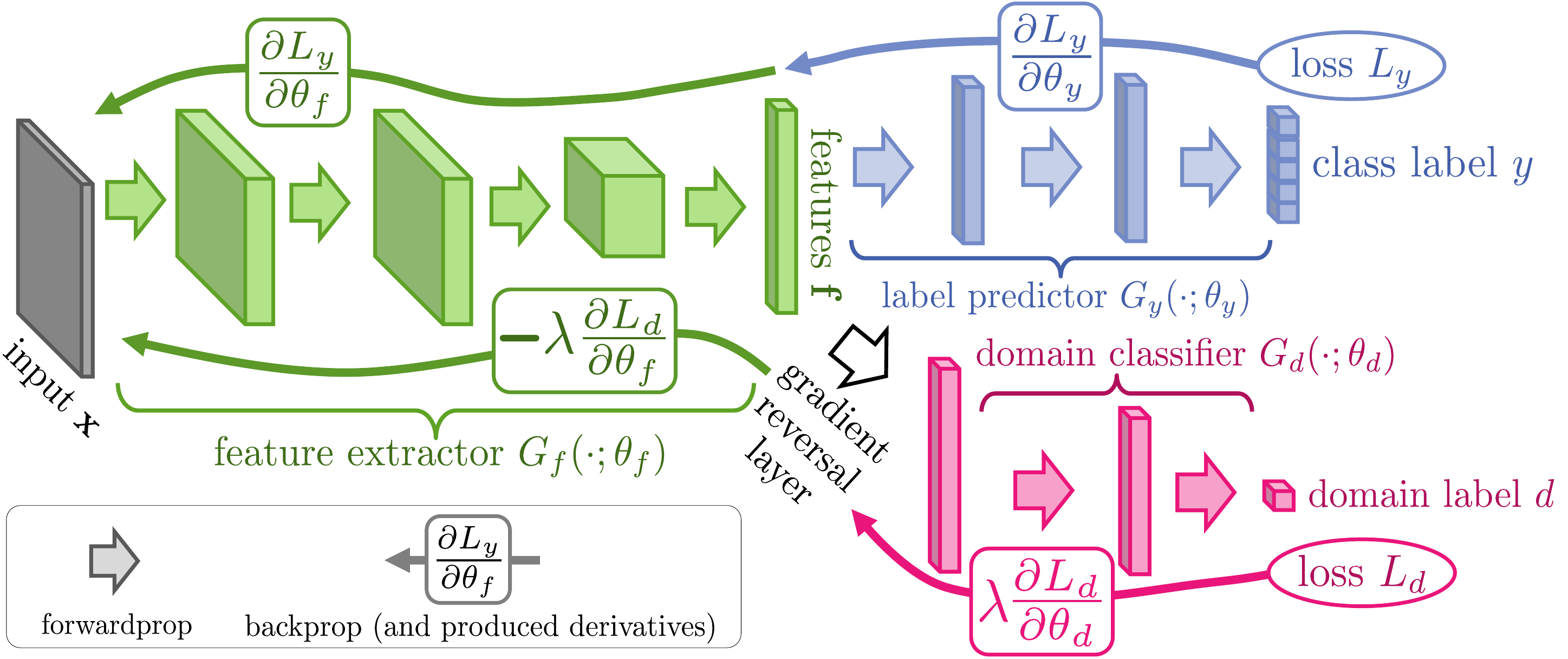}
    \caption{A visual representation of the DANN architecture \cite{DANN}}
    \label{fig:DANN}
\end{figure}

The domain-adversarial classifier is the noteworthy part of a DANN architecture, as shown in Figure~\ref{fig:DANN}.
It is characterized by two specific elements.
First, a $\lambda$ regularizer is applied to the domain-adversarial head.
This element is present to prevent the head from dominating the training of the network and completely preventing the classification head from learning.
Secondly, it provides some measure of control over the effect of this domain classifier over the total representation.
While a high $\lambda$ will learn a representation that will not distinguish the domains, it may not be the best classifier for the class labels.
In the reverse case, the domain classifier will not affect the total representation much, even if it would be helpful.
Thus, the $\lambda$ parameter is an additional hyper-parameter to tune for the algorithm.

The second element is the \emph{gradient reversal layer}.
This layer does not affect the forward propagation through the network, but will reverse (or multiply by $-1$) the gradient that comes back through it.
This is how the DANN architecture learns a representation that does not differentiate between domains, as the gradient goes in a direction where it does not reliably discern from which domain the examples come.

This general methodology can be applied to the current datasets, with minor changes.
First, instead of source and target domains we can consider the base domain and the stylized domain.
The style-transfer used to create the stylized domain from the base domain will conserve shape in the images, while changing or removing colour and texture clues.
Thus, a representation that cannot distinguish between the two domains should be reliant on shape more than other types of clues.
A further change applied to the DANN architecture was that all samples were used for the label predictors and for the domain classifier.
Thus, the experiments corresponded to the mixed experiments described above, but with the added element of the DANN domain classifier.

For all experimentations with DANN, the data augmentation was kept for the datasets as with the mixed experiments.
Moreover, the $\lambda$ parameter was tested in the range of $1^{-3}$ to $1^{-5}$, with the search space being expanded if the best results on the validation set were obtained on a boundary value.

\section{Results and Discussion}

\subsection{Training on Base Datasets}

Let us first consider the baseline case, where the models are trained on the base datasets.
These will form our baseline, comparing the robustness and accuracy of enhanced training methods.

\begin{table}[h]
    \centering
    \begin{tabular}{cccc}
        Dataset & Model & Base Accuracy & Stylized Accuracy \\ \hline
        Dice & ResNet & 99.94\% & 50.65\% \\
        Dice & SNet & 99.94\% & 40.88\% \\
        DvC & ResNet & 99.15\% & 87.42\% \\
        DvC & SNet & 97.73\% & 81.03\% \\
        DBI & ResNet & 94.58\% & 34.52\% \\
        DBI & SNet & 84.37\% & 22.07\% \\
        Food101 & ResNet & 92.90\% & 18.16\% \\
        Food101 & SNet & 90.09\% & 17.31\% \\
    \end{tabular}
    \caption{Baseline results of the models on the datasets. The reported accuracy corresponds to the Top-5 Accuracy in the case of the DBI and Food101 datasets. SNet denotes the SqueezeNet model.}
    \label{tab:base_res}
\end{table}

Table~\ref{tab:base_res} shows the performance of each model trained on the base data and tested on the test sets.
We can observe high accuracies in all datasets, with accuracies over 90\% on the base test set.
However, we can clearly remark the texture bias, as the accuracies are generally much lower on the stylized test sets.
While the models trained on the DvC dataset keep accuracies of over 80\% on their stylized test sets, the Dice dataset drops below 50\% accuracy (on 6 classes).

The DBI and Food101 datasets have their accuracies drop below 35\%, even 20\% in the case of Food101.
While each of the stylized accuracies are markedly better than random guessing, they are far below the base accuracy and show that the models are not adapted to predicting on stylized images.
Both the ResNet architecture and the SqueezeNet architecture show the same trends on the data, with the ResNet models having a higher accuracy overall.

\subsection{Training on Stylized Datasets}

\begin{table}[h]
    \centering
    \begin{tabular}{cccc}
        Dataset & Model & Base Accuracy & Stylized Accuracy \\ \hline
        Dice & ResNet & 99.76\% & 99.49\% \\
        Dice & SNet & 96.77\% & 97.84\% \\
        DvC & ResNet & 97.77\% & 96.00\% \\
        DvC & SNet & 95.45\% & 93.20\% \\
        DBI & ResNet & 85.54\% & 74.36\% \\
        DBI & SNet & 61.99\% & 47.35\% \\
        Food101 & ResNet & 77.00\% & 71.10\% \\
        Food101 & SNet & 71.68\% & 63.69\% \\
    \end{tabular}
    \caption{Results of the models trained on the stylized datasets. The reported accuracy corresponds to the Top-5 Accuracy in the case of the DBI and Food101 datasets. SNet denotes the SqueezeNet model.}
    \label{tab:style_res}
\end{table}

Table~\ref{tab:style_res} shows the results on the models trained on the stylized datasets.
Here, we see an overall decrease in base test set accuracy.
This is expected however, since the models were not trained on the base images.
The accuracy on the stylized test sets are however largely increased.
In the case of the Dice dataset, we can see accuracies over 97\%, equal or exceeding the base accuracy.
The stylized accuracies are also over 90\% for the models trained on the DvC dataset.

For the DBI and Food101 datasets, the stylized test set accuracies are more modest, between 45\% and 75\%.
They do still represent a significant increase from the models trained on the base dataset.
As mentioned in the datasets sections, the DBI and Food101 contain some classes that are, in reality, very similar to each other.
Thus, the relatively low accuracy when predicting on stylized images may simply be due to texture and colour cues that should be present, but have been removed by the style-transfer.

\subsection{Training on Mixed Datasets}

\begin{table}[h]
    \centering
    \begin{tabular}{cccc}
        Dataset & Model & Base Accuracy & Stylized Accuracy \\ \hline
        Dice & ResNet & 99.97\% & 99.76\% \\
        Dice & SNet & 99.97\% & 97.66\% \\
        DvC & ResNet & 99.07\% & 96.44\% \\
        DvC & SNet & 97.71\% & 93.49\% \\
        DBI & ResNet & 92.63\% & 75.70\% \\
        DBI & SNet & 83.87\% & 56.82\% \\
        Food101 & ResNet & 89.73\% & 73.31\% \\
        Food101 & SNet & 88.24\% & 65.72\% \\
    \end{tabular}
    \caption{Results of the models trained on the mixed datasets. The reported accuracy corresponds to the Top-5 Accuracy in the case of the DBI and Food101 datasets. SNet denotes the SqueezeNet model.}
    \label{tab:mixed_res}
\end{table}

\begin{figure*}[!htb]
    \centering
    \subfloat[]{
       \includegraphics[width=7cm]{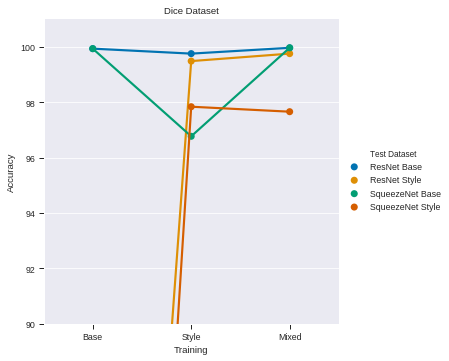}
    }
    \subfloat[]{
       \includegraphics[width=7cm]{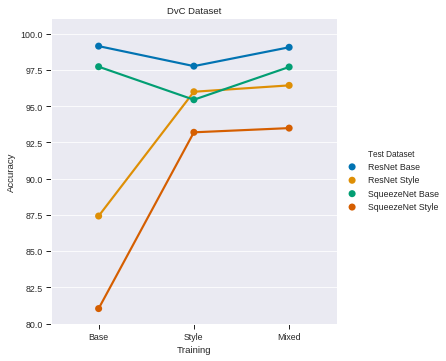}
    } \\
    \subfloat[]{
       \includegraphics[width=7cm]{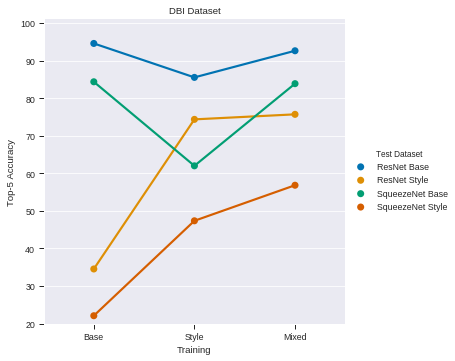}
    }
    \subfloat[]{
       \includegraphics[width=7cm]{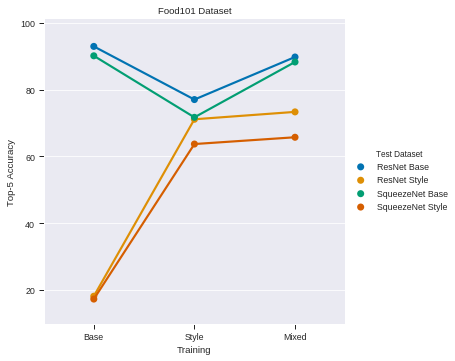}
    } 
    \caption{Results of base, stylized and mixed learning on the four datasets. For subfigures (a) and (b), we use the accuracy. For subfigures (c) and (d), the Top-5 accuracy is shown.}
    \label{fig:resgraph}
\end{figure*}

Table~\ref{tab:mixed_res} shows the performance of the models when trained on the mix of base and stylized data, as well as Figure~\ref{fig:resgraph} in graphical form.
For the Dice dataset, we can see similar, if slightly better (0.03\%), accuracy on the base testing set.
For the stylized test sets, there is a slight gain (0.3\%) for the ResNet architecture.
There is however no gain for the SqueezeNet architecture, compared with the training on stylized data.
Still, the mixed training yields equal or better performance on the base test set, and almost equal performance on the stylized test set.
For the DvC dataset, we see a similar effect.
For both ResNet and SqueezeNet models, the base accuracy is roughly equal (differences inferior to 0.1\%).
There are however very clear gains on the test set accuracy, even compared with the training on stylized datasets.
The ResNet model gains 0.44\% accuracy on stylized data, while the SqueezeNet model gains 0.29\%.
For these two datasets, the models trained on a mix of base and stylized data seem to have \emph{equal} accuracy and \emph{increased} robustness.

Let us now consider the last two datasets.
In the case of the DBI dataset, the ResNet model loses out on base accuracy (1.95\% loss), but the accuracy on the stylized dataset is once again better than when trained on stylized data (by 1.34\%).
We see a similar trend in the SqueezeNet model, with a slight decrease in accuracy compared to the base dataset (0.5\%).
The SqueezeNet model sees an even greater increase in stylized accuracy, with a gain of 9.47\%, compared to training with stylized data.
We see a greater loss in base test set accuracy in the Food101 dataset, at 3.17\% for the ResNet and 1.85\% for the SqueezeNet.
Both models do increase their stylized test set accuracy with the mixed learning however, respectively with increases of 2.21\% and 2.03\%.

For both these datasets, the models seem to \emph{decrease} in accuracy compared to models trained on the base data.
Yet, the representation learned by mixed training is much more \emph{robust}, with the stylized test set accuracy increasing compared to training on the stylized data.
The second effect was observed on all four datasets.
The conclusion in \cite{geirhos2018} indicated that this type of training should increase both accuracy and robustness.
The current results indeed show an increase in robustness and the models learning though shape-bias.
There is however no clear increase in accuracy for the models, compared with training on the base dataset.

\subsection{Using Domain-Adversarial Training}

\begin{table}[h]
    \centering
    \begin{tabular}{ccc}
        Dataset & Base Accuracy & Stylized Accuracy \\ \hline
        Dice & 99.97\% (0.00\%) & 99.70\% (-0.06\%) \\
        DvC & 98.95\% (-0.12\%) & 96.10\% (-0.34\%) \\
        DBI & 93.85\% (1.22\%) & 77.62\% (1.92\%) \\
        Food101 & 90.36\% (0.63\%) & 73.87\% (0.56\%) \\
    \end{tabular}
    \caption{Results of the ResNet models trained on the mixed datasets using a domain-adversarial network. The reported accuracy corresponds to the Top-5 Accuracy in the case of the DBI and Food101 datasets. The percentage in parentheses is the gain or loss of accuracy compared to mixed training.}
    \label{tab:da_res}
\end{table}

The Table~\ref{tab:da_res} shows the results of training the ResNet models with a domain-adversarial methods, on a mixed dataset of base and stylized data.
The accuracy on each test set is shown, as well as the difference between the domain-adversarial training and the mixed training of the networks.
Note that the results only include ResNet models, as the SqueezeNet models were not trained due to time and computational constraints.
For the Dice dataset, the results are equivalent, with a slight change of 0.06\% stylized accuracy.
This is however not statistically significant.
In the case of the DvC dataset, the domain-adversarial training causes a slight decrease in both base and stylized accuracies.

Let us now consider the DBI and Food101 datasets.
In both cases, domain-adversarial training of the network induces increases on the accuracies, for the base test set and the stylized test set.
This increase is greater than 1\% in both the base test set and stylized test set for the DBI dataset.
This results in the best performances for both datasets on the stylized test set.
This suggests that the domain-adversarial training did indeed help the networks learn a representation that is more robust and dependant upon shape.

Yet, the baseline test for the DBI and Food101 datasets still outperform the mixed training and the domain-adversarial models.
For the DBI dataset, the baseline was a base accuracy of 94.58\%, compared to 93.85\% using domain-adversarial training and style-transferred examples (a difference of 0.73\%).
Food101 has a best accuracy of 92.90\% in the baseline, compared to a 90.36\%, for a difference of 2.54\%.
The complete picture, over the different datasets, thus indicates that using a mix of base and stylized images does not increase the accuracy of a model, but will increase its robustness and shape-bias.
Moreover, domain-adversarial training over these base and stylized images seem to further increase the robustness of the model, and increase its accuracy over simply training on both types of images.

\section{Conclusion}

In conclusion, it is possible to increase the shape bias and robustness of a model for a given dataset using style-transfer and domain-adversarial training.
Over multiple datasets of varying complexity, we can see that models trained on a base dataset will have difficulty maintaining their performance if texture and colour clues are changed in the images.
Augmenting the dataset through a style-transfer will help the model learn a representation that can predict accurately on both the original dataset and on the stylized version.
While the results of \cite{geirhos2018} have not been replicated fully, as there is no clear gain in accuracy on the base dataset, the result is indeed more robust models.

Moreover, adding a domain-adversarial element to the training of the model further increases the robustness of the model and further encourage the representation learned to be dependant upon shape.
The domain-adversarial component also helps to increase the accuracy of the models, over the mixed training of base and stylized examples.

Further inquiries in this line of research are possible.
Firstly, \cite{geirhos2018} trained their ResNet architecture from scratch during their experiments on ImageNet.
Due to time and computational constraints, the models presented were pre-trained on ImageNet.
It is thus possible a new and possibly better representation could be found by training the models from scratch using domain-adversarial methods.
Secondly, the current work was performed on relatively small SqueezeNet 1.1 and ResNet-34 architectures.
It is possible that improvements could be made to the results using larger and higher performance architectures.

Finally, let us reflect on the significance of these results.
As was stated concerning some of the datasets, it is possible that the best possible answer to some problems indeed lie in the texture of an image, or its colour.
Thus, it could be better, for some problems, to train a model without using the proposed methodology.
It is also possible to envision the proposed methods as the basis of a type of data augmentation.
Indeed, while this would be very computationally intensive, it would be possible to style-transfer images at random during batches in order to augment shape-bias in training.
Moreover, the domain-adversarial element of the network provides regularization through its multi-task aspect \cite{evgeniou2004regularized}.
Even applied as was presented, we can see that, for some problems, the methodology provides added robustness and regularization with little to no decrease in accuracy.

\section*{Acknowledgements}

I wish to thank Nicolas Garneau and Frédérik Paradis for their counsel and for answering some questions about neural network design and implementation for this project. 
I also wish to thank François Laviolette and the Groupe de Recherche en Apprentissage Automatique de l'université Laval (GRAAL) research group for the use of computing resources. 
Finally, computations were made on the supercomputer Helios (Université Laval) and Graham (University of Waterloo), managed by Calcul Québec and Compute Canada. 
The operation of this supercomputer is funded by the Canada Foundation for Innovation (CFI), the ministère de l'Économie, de la science et de l'innovation du Québec (MESI) and the Fonds de recherche du Québec - Nature et technologies (FRQ-NT). 

\bibliography{bibliography}
\bibliographystyle{icml2019}

\end{document}